\documentclass[conference]{IEEEtran}
\IEEEoverridecommandlockouts

\pdfoutput=1

\usepackage{cite}
\usepackage{amsmath,amssymb,amsfonts}
\usepackage{algorithmic}
\usepackage{graphicx}
\usepackage{textcomp}
\usepackage{xcolor}

\usepackage[american]{babel}
\usepackage{microtype}
\usepackage{epstopdf}
\usepackage{multirow}
\usepackage{booktabs} 
\usepackage{float}
\usepackage{bm}
\usepackage{amssymb}
\usepackage{amsmath}
\usepackage{subfigure}
\usepackage{xcolor}
\usepackage{array}
\usepackage{color}

\newcommand{\redbold}[1]{\textcolor{red}{\textbf{#1}}}

\begin{document}

\title{Learning review representations from user and product level information for spam detection}

\author{
	\IEEEauthorblockN{Chunyuan Yuan, Wei Zhou\textsuperscript{*}, Qianwen Ma, Shangwen Lv, Jizhong Han, Songlin Hu}
	\IEEEauthorblockA{
		\textit{School of Cyber Security, University of Chinese Academy of Sciences} \\
		\textit{Institute of Information Engineering, Chinese Academy of Sciences}\\
		Beijing, China \\
		\{yuanchunyuan, zhouwei, maqianwen, lvshangwen, hanjizhong, husonglin\}@iie.ac.cn
	}
	\thanks{* is the corresponding author.}
}

\maketitle

\begin{abstract}
Opinion spam has become a widespread problem in social media, where hired spammers write deceptive reviews to promote or demote products to mislead the consumers for profit or fame. Existing works mainly focus on manually designing discrete textual or behavior features, which cannot capture complex semantics of reviews. Although recent works apply deep learning methods to learn review-level semantic features, their models ignore the impact of the user-level and product-level information on learning review semantics and the inherent user-review-product relationship information. 

In this paper, we propose a \textbf{H}ierarchical \textbf{F}usion \textbf{A}ttention \textbf{N}etwork (\textbf{HFAN}) to automatically learn the semantics of reviews from user and product level. Specifically, we design a multi-attention unit to extract user(product)-related review information. Then, we use orthogonal decomposition and fusion attention to learn a user, review, and product representation from the review information. Finally, we take the review as a relation between user and product entity and apply TransH to jointly encode this relationship into review representation. Experimental results obtained more than 10\% absolute precision improvement over the state-of-the-art performances on four real-world datasets, which show the effectiveness and versatility of the model. 
\end{abstract}

\begin{IEEEkeywords}
Opinion mining, Opinion spam detection, Hierarchical fusion attention network, User-level and product-level information
\end{IEEEkeywords}

\section{Introduction}
Online reviews play an important role for individuals and organizations when people make vote or purchase decisions. Considering these great benefits, many spammers have been employed to write deceptive reviews to influence users' decisions. The news from BBC has shown that nearly 25\% of Yelp reviews could be fake.\footnote{http://www.bbc.com/news/technology-24299742} Another piece of BBC news reports that Samsung hired spammers to write fake reviews on web forums.\footnote{http://www.bbc.com/news/technology-22166606} This spam case has been punished by the Fair Trade Commission in Taiwan, and researched by previous work~\cite{chen2015opiniona}. Reports like these are emerging in an endless stream. These spam reviews could significantly mislead consumers and damage the reputations of the websites. Therefore, it is urgent to propose some methods to automatically detect spams and make reviews more authentic.

Fortunately, many effective methods have been proposed. Nitin and Liu~\cite{jindal2008opinion} firstly put forward this problem and named it as opinion spam detection. Subsequent works are mainly dedicated to designing elaborate features to improve detection performance. For example, textual features like psychological and linguistic clues~\cite{ott2011finding}, syntactic stylometry~\cite{feng2012syntactic}, review topic~\cite{li2013topicspam,chen2019nonparametric}, and behavioral features like rating deviation~\cite{mukherjee2013yelp,rayana2015collective} are explored in many works. However, designing effective features is usually time-consuming~\cite{wang2016learning} and heavily rely on expert knowledge in particular areas. 


Motivated by the great success of deep learning in natural language process, recent methods~\cite{li2017document,ren2016deceptive,wang2017detecting} introduce deep learning models into spam detection, which shows neural network models can capture complex semantic information that is difficult to express using traditional discrete manual features~\cite{ren2016deceptive}. Despite the success of deep learning methods, they mainly utilize the review text information while ignoring the important influences of user- and product- level information on learning the review semantic representation and the inherent relationship among users, reviews, and products.


\begin{table}[!htbp]
	\centering
	\vspace*{-1\baselineskip}
	\caption{Examples of user level and product level integral characteristics.}
	\scriptsize
	\setlength{\tabcolsep}{0.1mm}{ 
		\begin{tabular}{|c|l|c|}
			\hline
			\textbf{user id} &    \makebox[7cm][c]{\textbf{spam review}} & \textbf{product id} \\ \hline
			\multirow{3}[2]{*}{u52} & \redbold{loveeeeee this place...} happy hour is \redbold{best!!} especially on tuesdays ... & p410 \\ \cline{2-3} 
			& \redbold{love this brunch...} cant believe that it is still great being that it .... & p454 \\ \cline{2-3} 
			& one of the best food places... \redbold{love this place...} simple simple simple...  & p81 \\  \cline{2-3} 
			& \redbold{Love this place for brunch}/happy hour\redbold{!!...} best time to go during ...  & p605 \\ \hline 
			
			u1394   & \redbold{great food}, \redbold{great service}. my wife and our friends love it. Can't ...  & \multirow{3}[2]{*}{p621} \\ \cline{1-2}
			u147   &  \redbold{Great food} and \redbold{service} was friendly and quick!  We tried this ... &  \\ \cline{1-2} 
			
			u1973   &  \redbold{Great food}. Definitely worth checking out. A little to busy ... & \\ \cline{1-2} 
			u2496   & \redbold{Great food} \redbold{excellent service} but quit noisy. & \\ \hline
		\end{tabular}
	}
	\label{running_example}
	\vspace*{-1\baselineskip}
\end{table}

To illustrate the problem, we choose several spam review examples from YelpNYC~\cite{rayana2015collective} dataset. Referring to Table~\ref{running_example}, every review seems normal from the single review level. However, if we see these reviews from user level ``u52'' or product level ``p621'', we can find some obvious abnormal patterns such as ``Love this place'', ``love this brunch'', ``Great food'', and ``Great service''. Actually, there are many different patterns from the user and product levels, such as emotional polarity, writing habit, or overall qualities of products. These user-level or product-level integral characteristics are hard to be displayed at the single review level, which presents great challenges to detect spam reviews only by review-level semantic information. 

Based on the above problems, we design a hierarchical fusion attention network to learn semantic representations from the user and product level and encode the user-review-product relationship. Firstly, we design a user(product)-related multi-attention unit respectively to extract the user(product)-related semantic features to form the sentence representation. Then, we apply orthogonal decomposition and fusion attention units to learn the user(product)-related information from the sentence representations as user (product) representation. Finally, we take reviews as relations between the user and product entities. Considering a user may post several reviews about one product or topic, so it is a one-to-many mapping relation, thus we use TransH~\cite{wang2014knowledge} to encode the user-review-product relationship.

In conclusion, our major contributions include:
\begin{itemize}
	\item We design multi-attention units and fusion attention units to facilitate learning semantic representations from user level and product level.
	\item We propose a unified network to jointly fuse the review and the user-review-product relationship for opinion spam detection.
	\item Experimental results on four public datasets achieve significant improvement, which shows the effectiveness and versatility of HFAN to detect spam reviews. 
\end{itemize}


\section{Related Work} \label{related_works_section}
\subsection{Feature-based Methods}
The opinion spam detection problem was firstly studied by~\cite{jindal2008opinion}. Their work demonstrated that opinion spams were widespread. Since then, opinion spam detection has been drawing increasing attention. \cite{ott2011finding} applied psychological and linguistic features to classify opinion spams. \cite{feng2012syntactic} investigated syntactic stylometry for deceptive detection. \cite{li2013topicspam} proposed a generative LDA-based topic modeling approach for fake review detection. \cite{chen2015opiniona} conducted a real case study based on a set of internal spam records leaked from a shady marketing campaign. They explored the characteristics of opinion spams and spammers in a web forum and used SVM model to detect spams. \cite{wang2016learning} explored to learn the review representation with behavioral information by tensor decomposition.

Feature-based methods use manually designed discrete features, which can be sparse and fail to effectively encode the semantic information~\cite{ren2016deceptive}. Moreover, designing effective features is usually time-consuming and heavily rely on expert knowledge in particular areas. In this paper, we propose a model to automatically learn semantic representations from raw review data for better detecting opinion spam.



\subsection{Deep Learning Methods}
Considering manual discrete features cannot encode the review semantics from the discourse perspective, Ren et al.~\cite{ren2016deceptive} firstly proposed a hierarchical model to learn the review representation from word and sentence level. Subsequently, Li et al.~\cite{li2017document} explored to combine these automatically learned review representation with traditional linguistic and syntax features. You et al.\cite{you2018attribute} embedding texts, behavior features, and entity attributes to solve the cold-start spam review detection problem.

Although the deep learning methods achieve good performances for spam detection, these models mainly utilize review text information and behavior information, which ignore the important influences of users and product level information on learning the review semantic representation and the user-review-product relationship. In this paper, we try to learn review representation from user level and product level and jointly embed the relationship into the review representation. In this way, our model can capture higher level information that is hard to be learned on the single review level.



\section{Proposed Model}  \label{model_section}
In this section, we propose a unified model for opinion spam detection. We will describe (1) how to automatically extract user(product)-related words or phrases from raw review text; (2) how to learn user, product and review representation from these review features; and (3) how to fuse the user-review-product relationship into the model.

\begin{figure*}[!htbp]
	\centering
	\includegraphics[scale=0.7]{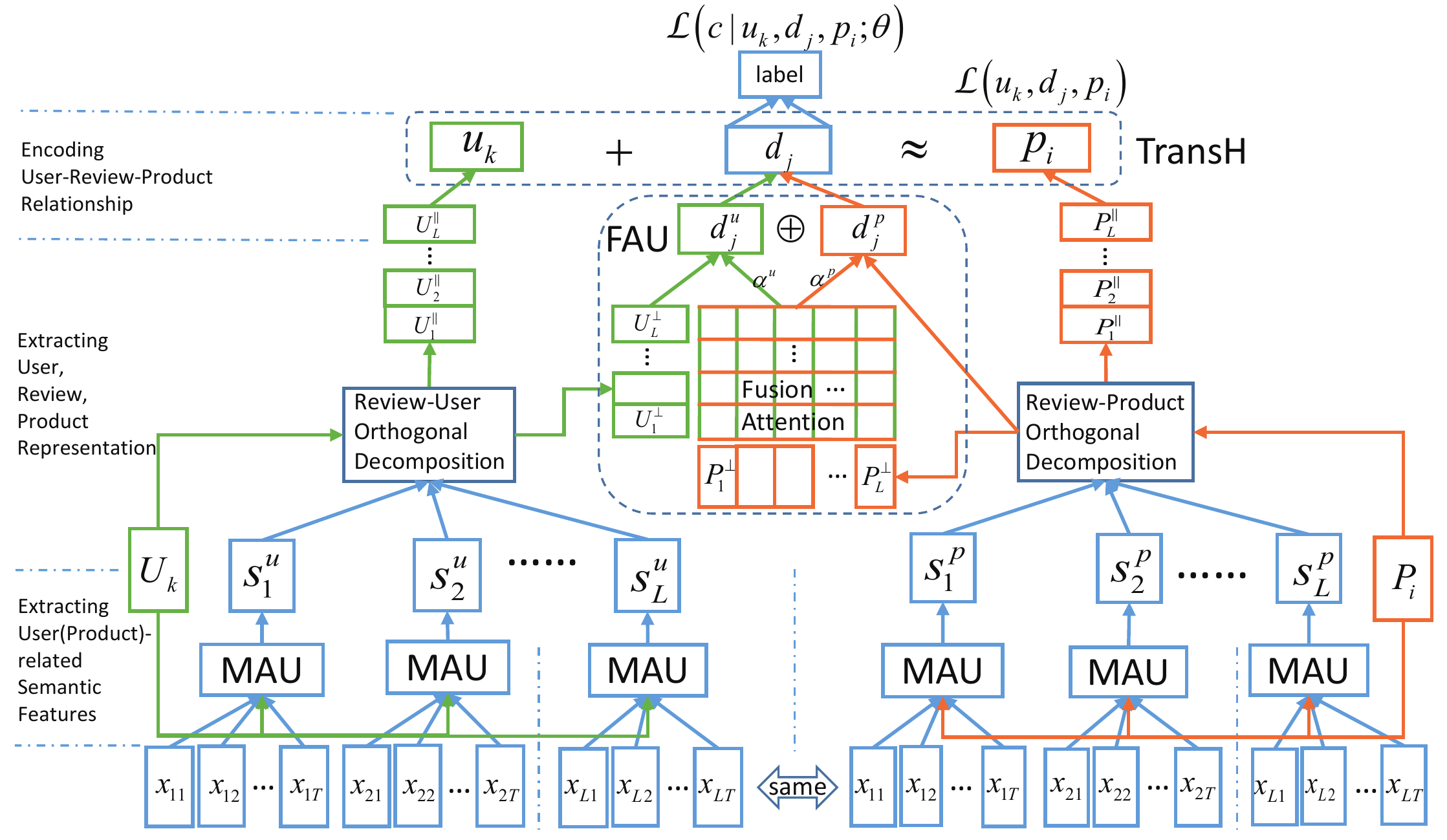}
	\caption{The HFAN model architecture. $U_k$ is the user who posts the review and $P_i$ is the product or topic that the review talks about. $U_k$ is shared by all reviews he posts. $P_i$ is shared by all reviews that talks about it.}
	\label{model}
	\vspace*{-1\baselineskip}
\end{figure*}

We first define some notations. Figure~\ref{model} shows the network architecture of HFAN. The inputs of the model are the review document $\bm{D}$, the product $\bm{P}$ that the review talks about, and the user $\bm{U}$ who posts the review. Each review has $L$ sentences and each sentence contains $T$ words. The outputs of the network are class probabilities. We use $p\left(c|\bm{U},\bm{D}, \bm{P}; \bm{\theta} \right)$ to represent the probability of the sample being class $c$, where $\bm{\theta}$ represents all the parameters in the network.

\subsection{Extracting User(Product)-related Semantic Features}
From user's view, not all words reflect a user's habit or preference. To capture the user-related semantic features of the review at word level, we design the multi-attention unit. 

%



\subsubsection{Multi-Attention Unit (\textbf{MAU})}
Firstly, the local context matrix of the word $x_j$ is represented as the concatenation of words' embedding $\bm{X} = \bm{x}_{j-r:j+r}$ $(0 \le j \le T)$, and $r$ is radius of the local region. Since words contribute to the sentence representation differently, we apply the attention mechanism to summarize the local context matrix to extract user-related words that are important to represent the meaning of the sentence:
\begin{equation}
\begin{split}
& \bm{v}^{(.)}_j = \sum_{t=1}^{2r+1} \alpha^{(.)}_t \bm{X} \,,  \\
& \alpha^{(.)}_t = \frac{\exp\left(\bm{u}^{(.)}_t \right)}{\sum_{k=1}^{2r+1} \exp\left(\bm{u}^{(.)}_k \right)} \,, \\
& \bm{u}^{(.)} = \tanh\left(\bm{X}\bm{W}^{(.)}_x + \bm{U}_j\bm{W}^{(.)}_u\right) \,,  
\end{split}
\end{equation}
where $\bm{v}^{(.)}_j \in \mathbb{X}^{1 \times d}$ and $\bm{\alpha^{(.)}}$ is score function which determines the importance of words for composing sentence representation about the current user. $\bm{W}^{(.)}_x, \bm{W}^{(.)}_u \in \mathbb{R}^{d \times d}$ are transformation matrices and $\bm{U}_j \in \mathbb{R}^{(2r+1) \times d}$ is $2r+1$ copies of the user embedding $\bm{u}_j$.

We use multiple attention units ($m$ units) to capture multiple representations from different semantic subspaces. Then, max pooling is applied to every feature dimension to select the most informative features:
\begin{equation}
\bm{v}_j = max\left( \left[\bm{v}^{(1)}_j, \bm{v}^{(2)}_j, \ldots, \bm{v}^{(m)}_j \right] \right) \,,
\end{equation}
which generates the local context representation of word $x_j$.

We concatenate the word embedding $\bm{x}_j$ with its user-related context representation $\bm{v}_j$ to compose the word representation. Then, all word representations are stacked to form the sentence matrix $\bm{V}_i = \left[\bm{x}_1 \oplus \bm{v}_1, \bm{x}_2 \oplus \bm{v}_2, \ldots, \bm{x}_T \oplus \bm{v}_T \right]$, where $\bm{V}_i \in \mathbb{R}^{T \times 2d} $.

To obtain the sentence representation, we use linear layer and max pooling on sentence matrix:
\begin{equation}
\begin{split}
& \bm{S}_i = \tanh\left(\bm{V}_i \bm{W}_v + \bm{b}\right) \,, \\
& \bm{s}^u_i = \max_{dim=1} \left(\bm{S}_i \right)
\end{split}
\end{equation}
where $\bm{W}_v \in \mathbb{R}^{2d \times d}$. In this way, the encoder can select most discriminative features to form the sentence representation $\bm{s}^u_i \in \mathbb{R}^{1 \times d}$ for sentence $i$. 

Similarly, not all words reflect product's quality and property. Therefore, we can generate product-related sentence representation $\bm{s}^p_i$ in the same way.

\subsection{Extracting User and Product Representation}
We have obtained user-related sentence matrix $[\bm{s}^u_1, \bm{s}^u_2, \ldots, \bm{s}^u_L]$ and product-related sentence matrix $[\bm{s}^p_1, \bm{s}^p_2, \ldots, \bm{s}^p_L]$. Then, we will distill user and product representation from the sentence matrices. 

To extract user representation, we apply orthogonal decomposition~\cite{wang2016sentence} to decompose the user-related sentence representation $\bm{s}_i^u, i \in \{1,2, \ldots, L\}$ to parallel and orthogonal direction of user embedding $\bm{U}_k$: 
\begin{equation}
\begin{split}
& \bm{U}^{\parallel}_i = \frac{\bm{s}_i^u \bm{U}_k^T}{\bm{U}_k\bm{U}_k^T} \bm{U}_k \,, \quad \bm{U}^{\bot}_i = \bm{s}_i^u - \bm{U}^{\parallel}_i \,, \\
\end{split}
\end{equation}
the parallel part of the decomposition represents the information that is related to the user, and the orthogonal part corresponds to the additional information provided by the reviews that is irrelevant to the user.

The user representation can be represented as $\bm{u}_k = \mathop{mean}(\bm{U}^{\parallel}) \in \mathbb{R}^{1 \times d}$, where $\bm{U}^{\parallel} = [\bm{U}^{\parallel}_1, \bm{U}^{\parallel}_2, \ldots, \bm{U}^{\parallel}_L] \in \mathbb{R}^{L \times d}$ is the sentence representation matrix that parallels to the user embedding. 

By the same way, we can obtain product representation $\bm{p}_i = \mathop{mean}(\bm{P}^{\parallel}) \in \mathbb{R}^{1 \times d}$ and the additional information that is irrelevant to the product $\bm{P}^{\bot}$.

\subsection{Extracting Review Representation}
After above procedures, we have obtained the sentence representation matrix $\bm{U}^{\bot}$ and $\bm{P}^{\bot}$. Both representations contain different information of reviews from different views, thus we will explore to fuse two representation matrices to obtain a better review representation. Furthermore, we explore how to effectively incorporate the inherent user-review-product relationship into the review representation.

\subsubsection{Fusion Attention Unit(\textbf{FAU})} 
Referring to Figure~\ref{model}, two linear transformation matrices are respectively applied on two representations $\bm{U}^{\bot}$ and $\bm{P}^{\bot}$ to transform them into the semantic space:
\begin{equation}
\begin{split}
& \bm{D}^{u} =  \bm{U}^{\bot}\bm{W}_u  \,,  \quad  \bm{D}^{p} =  \bm{P}^{\bot}\bm{W}_p  \,,   \\
\end{split}
\end{equation}
where $\bm{W}_u, \bm{W}_p \in \mathbb{R}^{d \times d}$. Then, gated mechanism is applied to cross activate each other: 
\begin{equation}
\begin{split}
&\hat{\bm{D}}^{u}  = \bm{D}^{u} \odot \sigma \left(\bm{D}^{p} \right) \,, \quad  \hat{\bm{D}}^{p}  = \bm{D}^{p} \odot \sigma \left(\bm{D}^{u} \right) \,. \\
\end{split}
\label{ura_fomulation1}
\end{equation}
We treat the $\bm{D}^{p}$ as the gate to control how much review information $\bm{D}^{u}$ is allowed to flow to the next layer. Similarly, we use the review information $\bm{D}^{u}$ to control how much the $\bm{D}^{p}$ to impact the following layer.

After that, we use dot-product attention to compute the fusion matrix to build the connection between the two reviews matrices:
\begin{equation}
\bm{M} = \tanh \left(\bm{\hat{D}^{u}} \bm{\hat{D^p}}^T \right)    \,,
\label{ura_fomulation2}
\end{equation}
where $\bm{M} \in \mathbb{R}^{L \times L}$, every element $\bm{M}_{ij}$ represents the pair-wise correlation score of two review representations $\bm{\hat{D}_i^u}$ and $\bm{\hat{D}_j^p}$. 

Then, mean pooling operation is performed to average the correlation score over rows and columns of $\bm{M}$ respectively.  And $\mathop{softmax}(\bm{x_i}) = \frac{\exp{\bm{x}_i}}{\sum_{j} \exp{\bm{x}_j}}$ function is applied to normalize the correlation scores to get the attention weights:
\begin{equation}
\begin{split}
& \bm{\alpha}^u = \mathop{softmax} \left(\mathop{mean}_{dim=1}\left(\bm{M}\right)\right)  \,, \\
& \bm{\alpha}^p = \mathop{softmax} \left(\mathop{mean}_{dim=2}\left(\bm{M}\right)\right)  \,,
\label{ura_fomulation3}
\end{split}
\end{equation}
where $\bm{\alpha}^u, \bm{\alpha}^p \in \mathbb{R}^{1 \times L}$.

Finally, we get another two review representations by applying the attention weights on their original feature vectors:
\begin{equation}
\begin{split}
& \bm{d}^u_j = \bm{\alpha}^u \bm{D}^{u}  \,,  \quad  \bm{d}^p_j = \bm{\alpha}^p \bm{D}^{p}  \,, \\
\end{split}
\label{ura_fomulation4}
\end{equation}
where $\bm{d}^u_j, \bm{d}^p_j \in \mathbb{R}^{1 \times d}$. Since both representations are high-level representations of review from different views, we concatenate them as the final review representation:
\begin{equation}
\begin{split}
& \bm{d}_j = \bm{W} \left( [\bm{d}^u_j ; \bm{d}^p_j] \right)  \,, 
\end{split}
\label{ura_fomulation5}
\end{equation}
where $\bm{W} \in \mathbb{R}^{2d \times d}$ is a transformation matrix.

\subsection{Encoding User-Review-Product Relationship Information} 
The inherent relationship of users, reviews, and products contain abundant interactive information among users about products, which can reflect the closeness among users. We treat the user $\bm{u}_k$ and product $\bm{p}_i$ as head and tail entity respectively and take the review $\bm{d}_j$ as a relation between them. A user usually posts many reviews about the same product, so it is a one-to-many relationship between users and products. Thus, we propose to apply TransH~\cite{wang2014knowledge} to model this kind of relationship. 

We treat the inherent relationship (users, reviews, products) as a kind of regularization, which can help to build robust review representations. On the one hand, we hope that the similarity distance between users and their reviews is as close as possible. On the other hand, we also hope that the distance among user and reviews posted by other users is as far as possible. More formally, the relation loss can be described as:

\begin{equation}
\begin{split}
& \mathcal{L}(\bm{u}_k, \bm{d}_j, \bm{p}_i) = \frac{1}{|\Delta^{\prime}|} \! \sum_{(\bm{u}^{\prime}, \bm{d}^{\prime}, \bm{p}^{\prime}) \in \Delta^{\prime} } \! \max (0, L)  \,,  \\
&  L = l(\bm{u}_k, \bm{d}_j, \bm{p}_i) - l(\bm{u}^{\prime}, \bm{d}^{\prime}, \bm{p}^{\prime}) + 1  \,. 
\end{split}
\label{transh}
\end{equation}
where $\Delta^{\prime}$ denotes the set of negative triplets whose heads or tails are randomly replaced by other entities. Other constraints are the same as TransH~\cite{wang2014knowledge}. The distance function is defined as: 
\begin{equation}
\begin{split}
& l(\bm{u}, \bm{d}, \bm{p}) = || (\bm{u} - \bm{w}_d^T \bm{u} \bm{w}_d) + \bm{d} - (\bm{p} - \bm{w}_d^T \bm{p} \bm{w}_d)  ||^2_2   \,,  \\
\end{split}
\label{distance_fun}
\end{equation}
where $\bm{w}_d$ is the relation-specific hyperplane. We project the user and product entity to the hyperplane, and it enables different roles of an entity in different relations/triplets. In this way, we can solve the one-to-many problem.




\subsection{Opinion Spam Classification}
We use review representation $\bm{d}_j$ as features to detect spam review. The fully connected layers are applied over $\bm{d}_j$, and $softmax(\cdot)$ function is used to convert the output numbers into probabilities:
\begin{equation}
\begin{split}
& \bm{y} = \bm{W}_c(relu\left( \bm{d}_j\bm{W}_d + \bm{b}_d \right)) \,, \\
& p_i\left(c|\bm{u}_k, \bm{d}_j, \bm{p}_i ; \theta\right) = \frac{\exp\left(\bm{y}_i\right)}{\sum_{k=1}^{c}\exp\left(\bm{y}_k\right)} \,,
\end{split}
\end{equation}
where $\bm{W}_c, \bm{W}_d \in \mathbb{R}^{d \times d}$ are transformation matrix.

\subsubsection{Classification Loss}
Similar to previous works~\cite{ren2016deceptive,li2017document,wang2017detecting}, we use the cross entropy loss as the objective function to optimize the classification task: 
\begin{equation}
\mathcal{L}\left(c|\bm{u}_k, \bm{d}_j, \bm{p}_i; \theta \right) = -\sum_{i}\log p_i\left(c|\bm{u}_k, \bm{d}_j, \bm{p}_i ; \theta\right) \,.
\end{equation}

\subsubsection{Overall Loss and Optimization}
The overall loss of our model is the weighted sum of classification loss and the relation loss:
\begin{equation}
\begin{split}
& \mathcal{L}(\theta) \! = \! \sum_{i} \mathcal{L}\left(c|\bm{u}_k, \bm{d}_j, \bm{p}_i; \theta \right) \! + \! \beta \! \sum_{k} \! \sum_{i \in N\left(u_k\right)} \! \mathcal{L}\left(\bm{u}_k, \bm{d}_j, \bm{p}_i\right)  \,, \\
\end{split}
\end{equation}
where $\beta$ is a hyper-parameter and will be tuned on the validation dataset. 


\section{Experiments} \label{expriments_section}

\subsection{Datasets}
We evaluate the effectiveness of HFAN on four spam datasets. The statistics of the datasets are shown in Table~\ref{dataset_statistics}.

\begin{table*}[!htbp]
	\centering
	\caption{Dataset statistics.}
	\setlength{\tabcolsep}{3mm}{ 
		\begin{tabular}{|l|rr|rr|rr|rr|rr|}
			\hline
			\multirow{2}{*}{Dataset}  &\multicolumn{2}{c|}{Mobile01\_FirstPost}  &\multicolumn{2}{c|}{Mobile01\_Reply} &\multicolumn{2}{c|}{YelpChi}  &\multicolumn{2}{c|}{YelpNYC}  &\multicolumn{2}{c|}{YelpZip} \\ 
			& training &  test     & training  &  test  & training  &  test & training  &  test  & training  &  test \\
			\hline
			Average \#words    &   203&  146&  68&  63&  165& 166& 138&    138 &   137& 137  \\
			\hline
			\#Spam Reviews     &   546&   208&    1,337&    1,020&   7,135& 1,784& 29,508&    73,77&  64,372& 16,094  \\
			\hline
			\#Non-spam Reviews &10,405& 5,662&  147,504&   66,005&  46,780& 11,696& 257,733&  64,434&  422,505 & 105,627 \\
			\hline
			\#Users            &   5,130&  3,520&  16,272&  12,310&   32,475& 10,856& 138,185&     49,355&   224,548& 81,855   \\
			\hline
		\end{tabular}
	}
	\label{dataset_statistics}
	\vspace*{-1.5\baselineskip}
\end{table*}


\noindent\textbf{Mobile01 Review}\footnote{http://nlg3.csie.ntu.edu.tw/m01-corpus} is obtained from paper~\cite{chen2015opiniona}, which contains two subsets: first post subset (Mobile01\_FirstPost) and reply subset (Mobile01\_Reply). This dataset contains a set of internal records of opinion spams leaked from a shady marketing campaign reported by BBC~\footnote{http://www.bbc.com/news/technology-22166606}. 


\noindent\textbf{YelpChi}, \noindent\textbf{YelpNYC} and \textbf{YelpZip}\footnote{http://shebuti.com/collective-opinion-spam-detection} datasets are obtained from paper~\cite{rayana2015collective}, which are three public spam detection datasets crawled from the Yelp website. 



\subsection{Baselines}
To illustrate the effectiveness of HFAN, we select several state-of-the-art methods for comparison, including conventional feature-based methods and some recently proposed deep learning models for the spam detection.

\subsubsection{Feature-based Methods}

\

\noindent\textbf{SVM + Bag of Words (BoW)/n-grams + BF} mainly use machine learning algorithms with unigram, bigram, trigram. \textbf{Behavior Features (BF)} are obtained from papers~\cite{chen2015opiniona,mukherjee2013yelp,rayana2015collective}. 



\noindent\textbf{RSD}~\cite{wang2011review} is an iterative model to quantify the trustiness of reviewers, the honesty of reviews, and the reliability of stores. They propose a heterogeneous graph model to capture spamming clues.


\noindent\textbf{SpEagle}~\cite{rayana2015collective} is a graph-based method that combines linguistic features, behavioral features and reviews graph structure features and utilizes the Loopy Belief Propagation algorithm to compute the belief scores for reviews and users. 

\noindent\textbf{TDSD}~\cite{wang2016learning} is a tensor decomposition model to automatically learn the review representation and users' behavior information. They extended 11 interactive relations to embed the reviewers and products, which are concatenated with the review representation for spam detection.

\noindent\textbf{CHMM}~\cite{li2017bimodal} is the Coupled Hidden Markov Model (HMM) model with two parallel HMMs that incorporate both the reviewer's posting behavior and co-bursting behaviors from other reviewers.

\noindent\textbf{Spam2Vec}~\cite{maity2018spam2vec} is a framework to collectively use both review content and network information for spam detection. 


\subsubsection{Deep Learning Methods}

\

\noindent\textbf{CNN-GRNN}~\cite{ren2016deceptive} is the first model designed for spam detection. Convolutional neural network and gated recurrent network are applied on word level and sentence level respectively to learn the discourse information of reviews.


\noindent\textbf{SWNN}~\cite{li2017document} utilizes CNN to extract local semantic features, and apply KL-divergence to obtain the importance weight of the word. Then, a weight pooling is applied to transform sentence vectors into a document vector. 


\noindent\textbf{ABNN}~\cite{wang2017detecting} is an attention-based network by jointly embedding linguistic and behavioral features for spam detection. 


\noindent\textbf{AEDA}~\cite{you2018attribute} is a deep learning architecture for incorporating entities and their
inherent attributes from various domains into a unified framework.

\subsection{Experiment Settings}
We employ the same evaluation metrics and preprocessing procedures used in previous works~\cite{chen2015opiniona,rayana2015collective} in the experiments. Specifically, for the Mobile01 dataset, precision (P), recall (R), and F-measure ($F_1$) are used as evaluation metrics. For YelpChi, YelpNYC and YelpZip dataset, average precision (AP) and area under the curve (AUC) are used as evaluation metrics.


\subsubsection{Data Preprocessing} 
Most of preprocessing procedure is identical to previous works, different parts are listed as follows. For hierarchical models (including baselines and our model), we segment the document into sentences and words. As the average \#words of Mobile01\_FirstPost dataset is 203 (cf. Table~\ref{dataset_statistics}), so we set the maximum text length to 500, and any reviews longer than 500 will be truncated to 500. The maximum length of reviews in Mobile01\_Reply dataset is set to 100. For YelpChi, YelpNYC and YelpZip dataset, the maximum length is set to 200.

\subsubsection{Model Training} 
The dimension of the pre-trained embeddings is set to 300. In the experiments, we perform 3 folds cross-validation on the region radius $r \in \{1, 2, \ldots, 10\}$ and loss weight $\beta \in \{0.01, 0.1, 1, 10, 100\}$ and choose those parameters that achieves best performance. The MAU size $m$ is set to 2. The Adadelta~\cite{zeiler2012adadelta} algorithm is applied to optimize overall loss. The learning rate is initialized as 1.0 and gradually decreased during training.





\subsection{Results and Analysis}
The experimental results are shown in Table~\ref{exp_results_on_samsung},~\ref{exp_results_on_yelp}. From the results, we can observe that:


\begin{table}[!htbp]
	\centering
	\vspace*{-1\baselineskip}
	\setlength{\tabcolsep}{0.7mm}{
		\caption{Experimental results on Mobile01 dataset. The state-of-the-art results have been highlighted by the underline.}
		\begin{tabular}{p{4.1cm}|ccc|ccc}
			\toprule
			\multirow{2}[3]{*}{Models} & \multicolumn{3}{c|}{Mobile01\_FirstPost} & \multicolumn{3}{c}{Mobile01\_Reply} \\
			& P& R& $F_1$ & P& R& $F_1$  \\
			\midrule
			M1: SVM + content + title (BoW)       &59.12 & 51.44 & 55.01  &15.60  &26.47  &19.63\\
			M2: M1 + time + thread                &72.37 & 52.88 & 61.11 &19.66  &30.98  &24.06 \\
			M3: M2 + sentiment on brands          &70.97 & 52.88 & 60.61 &25.59  &29.61  &27.45 \\
			TDSD + BF   			              &\underline{73.12} & 54.45 & 62.42 &26.31  &30.38  &28.20  \\
			CHMM                                  &68.51 & 54.58 & 60.76 &21.14  &31.44  &25.28  \\
			Spam2Vec                              &68.64 & 55.35 & 61.28 &26.35  &30.59  &28.31  \\
			\midrule
			CNN-GRNN                         &63.21  &64.42&  63.81&  23.42&  33.43& 27.54 \\
			SWNN                             &65.57  &57.69&  61.38&  22.53&  35.78& 27.65 \\
			ABNN                             &61.21  &\underline{68.27}&  64.55&  27.13&  35.00& 30.57 \\
			AEDA                                  &68.54 & 62.39 & \underline{65.32} & \underline{28.37}  &37.11  &\underline{32.16}  \\
			\midrule
			HFAN                        &\textbf{86.96}  &67.31  &\textbf{75.88}   &\textbf{61.17}   &\textbf{40.00}   &\textbf{48.37} \\
			\bottomrule
		\end{tabular}
		\label{exp_results_on_samsung}
	}
	\vspace*{-1\baselineskip}
\end{table}

\begin{table}[!htbp]
	\centering
	\setlength{\tabcolsep}{2mm}{
		\caption{Experimental results on YelpChi, YelpNYC and YelpZip datasets. }
		\begin{tabular}{p{2cm}|cc|cc|cc}
			\toprule
			\multirow{2}[2]{*}{Models} & \multicolumn{2}{c|}{YelpChi} & \multicolumn{2}{c|}{YelpNYC} & \multicolumn{2}{c}{YelpZip} \\
			& AP& AUC & AP& AUC & AP& AUC  \\
			\midrule
			RSD     				&15.18  &50.62 	&12.55	&54.15    &18.03  &59.82 \\
			SpEagle 				&32.36  &78.87	&27.57	&78.29    &35.45  &80.40 \\
			TDSD					&34.68  &78.82	&\underline{36.62}	&78.86    &45.15  &81.63 \\
			CHMM                    &35.14  &78.68  & 35.13 &78.71  &\underline{49.56}  &82.64  \\
			Spam2Vec                &34.25  &78.61  & 35.04 &78.35  &46.33  &81.21  \\
			\midrule
			CNN-GRNN                &35.02 &78.68      &35.47    &\underline{79.04}      &48.57   &\underline{81.87}    \\
			SWNN                    &34.13  &78.57   &34.79    &78.57   &46.79   &81.25 \\
			ABNN                    &34.48  &78.53   &35.80    &78.83   &48.19   &80.82 \\
			AEDA                    &\underline{36.76}  &\underline{79.14}   &35.13    &78.92   &48.52   &81.32  \\
			\midrule
			HFAN            &\textbf{48.87} &\textbf{83.24}  &\textbf{53.82}    &\textbf{84.78}   &\textbf{62.35}  &\textbf{87.28} \\
			\bottomrule
		\end{tabular}
		\label{exp_results_on_yelp}
	}
	\vspace*{-1\baselineskip}
\end{table}

(1) Most neural network based models outperform feature-based methods, but the improvement mainly comes from recall rather than precision. The improvement of performances further proves the idea that the neural network can capture more complex semantic information that is difficult to express using traditional discrete manual features~\cite{ren2016deceptive}. However, we notice that performance improvement mainly comes from recall rather than precision, which shows the review-level semantic information may not fully reflect the discrepancy of spam and non-spam.

(2) Our proposed model outperforms neural network based models and features based methods, and the spam detection precisions obtain significant improvement. Referring to Table~\ref{exp_results_on_samsung},~\ref{exp_results_on_yelp}, HFAN gets 10.56\%, 16.21\% F1 gains over the state of the art on Mobile01 FirstPost and Reply dataset respectively. On the YelpChi, YelpNYC and YelpZip datasets, HFAN outperforms the best performances by 4.1\%, 5.74\% and 5.41\% AUC score respectively. It is worth noting that the spam detection precisions get over 10\% absolute improvement on four datasets, which shows the product-level and user-level information and the user-review-product relationship information are critical to learning the differences between spam and non-spam reviews.



In conclusion, HFAN significantly and consistently outperforms the state-of-the-art models. The experiments on four datasets demonstrate that HFAN can comprehensively leverage product-level and user-level information and the user-review-product relationship, thus achieving good generalization performance in the spam detection task.

\section{Conclusion} \label{conclusion_section}
In this paper, we design a hierarchical fusion attention network that can learn semantic representations of reviews from the user and product level. Additionally, we encode the inherent relationship among users, reviews, and products into the model. To evaluate the performance of HFAN, we conduct a series of experiments on four public datasets. Compared with the state of the art, HFAN achieves significant improvement, which proves the effectiveness of the model.



\section{Acknowledge}
This research is supported in part by the Beijing Municipal Science and Technology Project under Grant Z191100007119008 and Z181100002718004, the National Key Research and Development Program of China under Grant 2018YFC0806900 and 2017YFB1010000.


\bibliographystyle{IEEEtran}
\bibliography{mx}

\begin{thebibliography}{10}
\providecommand{\url}[1]{#1}
\csname url@samestyle\endcsname
\providecommand{\newblock}{\relax}
\providecommand{\bibinfo}[2]{#2}
\providecommand{\BIBentrySTDinterwordspacing}{\spaceskip=0pt\relax}
\providecommand{\BIBentryALTinterwordstretchfactor}{4}
\providecommand{\BIBentryALTinterwordspacing}{\spaceskip=\fontdimen2\font plus
\BIBentryALTinterwordstretchfactor\fontdimen3\font minus
  \fontdimen4\font\relax}
\providecommand{\BIBforeignlanguage}[2]{{%
\expandafter\ifx\csname l@#1\endcsname\relax
\typeout{** WARNING: IEEEtran.bst: No hyphenation pattern has been}%
\typeout{** loaded for the language `#1'. Using the pattern for}%
\typeout{** the default language instead.}%
\else
\language=\csname l@#1\endcsname
\fi
#2}}
\providecommand{\BIBdecl}{\relax}
\BIBdecl

\bibitem{chen2015opiniona}
Y.-R. Chen and H.-H. Chen, ``Opinion spam detection in web forum: a real case
  study,'' in \emph{Proceedings of the 24th International Conference on World
  Wide Web}.\hskip 1em plus 0.5em minus 0.4em\relax International World Wide
  Web Conferences Steering Committee, 2015, pp. 173--183.

\bibitem{jindal2008opinion}
N.~Jindal and B.~Liu, ``Opinion spam and analysis,'' in \emph{Proceedings of
  the 2008 International Conference on Web Search and Data Mining}.\hskip 1em
  plus 0.5em minus 0.4em\relax ACM, 2008, pp. 219--230.

\bibitem{ott2011finding}
M.~Ott, Y.~Choi, C.~Cardie, and J.~T. Hancock, ``Finding deceptive opinion spam
  by any stretch of the imagination,'' in \emph{Proceedings of the 49th Annual
  Meeting of the Association for Computational Linguistics: Human Language
  Technologies-Volume 1}.\hskip 1em plus 0.5em minus 0.4em\relax Association
  for Computational Linguistics, 2011, pp. 309--319.

\bibitem{feng2012syntactic}
S.~Feng, R.~Banerjee, and Y.~Choi, ``Syntactic stylometry for deception
  detection,'' in \emph{Proceedings of the 50th Annual Meeting of the
  Association for Computational Linguistics: Short Papers-Volume 2}.\hskip 1em
  plus 0.5em minus 0.4em\relax Association for Computational Linguistics, 2012,
  pp. 171--175.

\bibitem{li2013topicspam}
J.~Li, C.~Cardie, and S.~Li, ``Topicspam: a topic-model based approach for spam
  detection,'' in \emph{Proceedings of the 51st Annual Meeting of the
  Association for Computational Linguistics (Volume 2: Short Papers)}, 2013,
  pp. 217--221.

\bibitem{chen2019nonparametric}
J.~Chen, Z.~Gong, and W.~Liu, ``A nonparametric model for online topic
  discovery with word embeddings,'' \emph{Information Sciences}, vol. 504, pp.
  32--47, 2019.

\bibitem{mukherjee2013yelp}
A.~Mukherjee, V.~Venkataraman, B.~Liu, and N.~S. Glance, ``What yelp fake
  review filter might be doing?'' in \emph{ICWSM}, 2013.

\bibitem{rayana2015collective}
S.~Rayana and L.~Akoglu, ``Collective opinion spam detection: Bridging review
  networks and metadata,'' in \emph{Proceedings of the 21th acm sigkdd
  international conference on knowledge discovery and data mining}.\hskip 1em
  plus 0.5em minus 0.4em\relax ACM, 2015, pp. 985--994.

\bibitem{wang2016learning}
X.~Wang, K.~Liu, S.~He, and J.~Zhao, ``Learning to represent review with tensor
  decomposition for spam detection,'' in \emph{Proceedings of the 2016
  Conference on Empirical Methods in Natural Language Processing}, 2016, pp.
  866--875.

\bibitem{li2017document}
L.~Li, B.~Qin, W.~Ren, and T.~Liu, ``Document representation and feature
  combination for deceptive spam review detection,'' \emph{Neurocomputing},
  vol. 254, pp. 33--41, 2017.

\bibitem{ren2016deceptive}
Y.~Ren and Y.~Zhang, ``Deceptive opinion spam detection using neural network,''
  in \emph{Proceedings of COLING 2016, the 26th International Conference on
  Computational Linguistics: Technical Papers}, 2016, pp. 140--150.

\bibitem{wang2017detecting}
X.~Wang, K.~Liu, and J.~Zhao, ``Detecting deceptive review spam via
  attention-based neural networks,'' in \emph{National CCF Conference on
  Natural Language Processing and Chinese Computing}.\hskip 1em plus 0.5em
  minus 0.4em\relax Springer, 2017, pp. 866--876.

\bibitem{wang2014knowledge}
Z.~Wang, J.~Zhang, J.~Feng, and Z.~Chen, ``Knowledge graph embedding by
  translating on hyperplanes,'' in \emph{Twenty-Eighth AAAI conference on
  artificial intelligence}, 2014.

\bibitem{you2018attribute}
Z.~You, T.~Qian, and B.~Liu, ``An attribute enhanced domain adaptive model for
  cold-start spam review detection,'' in \emph{Proceedings of the 27th
  International Conference on Computational Linguistics}, 2018, pp. 1884--1895.

\bibitem{wang2016sentence}
Z.~Wang, H.~Mi, and A.~Ittycheriah, ``Sentence similarity learning by lexical
  decomposition and composition,'' in \emph{Proceedings of COLING 2016, the
  26th International Conference on Computational Linguistics: Technical
  Papers}, 2016, pp. 1340--1349.

\bibitem{wang2011review}
G.~Wang, S.~Xie, B.~Liu, and S.~Y. Philip, ``Review graph based online store
  review spammer detection,'' in \emph{2011 IEEE 11th International Conference
  on Data Mining}.\hskip 1em plus 0.5em minus 0.4em\relax IEEE, 2011, pp.
  1242--1247.

\bibitem{li2017bimodal}
H.~Li, G.~Fei, S.~Wang, B.~Liu, W.~Shao, A.~Mukherjee, and J.~Shao, ``Bimodal
  distribution and co-bursting in review spam detection,'' in \emph{Proceedings
  of the 26th International Conference on World Wide Web}.\hskip 1em plus 0.5em
  minus 0.4em\relax International World Wide Web Conferences Steering
  Committee, 2017, pp. 1063--1072.

\bibitem{maity2018spam2vec}
S.~K. Maity, S.~KC, and A.~Mukherjee, ``Spam2vec: Learning biased embeddings
  for spam detection in twitter,'' in \emph{Companion Proceedings of the The
  Web Conference 2018}.\hskip 1em plus 0.5em minus 0.4em\relax International
  World Wide Web Conferences Steering Committee, 2018, pp. 63--64.

\bibitem{zeiler2012adadelta}
M.~D. Zeiler, ``Adadelta: an adaptive learning rate method,'' \emph{arXiv
  preprint arXiv:1212.5701}, 2012.

\end{thebibliography}
\end{document}